\documentclass[10pt,twocolumn,letterpaper]{article}

\usepackage{cvpr}
\usepackage{times}
\usepackage{epsfig}
\usepackage{graphicx}
\usepackage{amsmath}
\usepackage{amssymb}
\usepackage{bm}
\usepackage{booktabs}
\usepackage{color}
\usepackage[ruled]{algorithm2e}

\usepackage{caption}
\usepackage{subcaption}
\usepackage{tabularx}
\newcolumntype{C}[1]{>{\centering\arraybackslash}p{#1}}
\usepackage{multicol}

\usepackage[pagebackref=true,breaklinks=true,letterpaper=true,colorlinks,bookmarks=false]{hyperref}

\cvprfinalcopy 

\def\cvprPaperID{793} 

\ifcvprfinal\fi
\begin{document}

\title{PointAugment: an Auto-Augmentation Framework\\ for Point Cloud Classification}


\author{Ruihui Li \quad Xianzhi Li \quad Pheng-Ann Heng \quad Chi-Wing Fu
    \vspace{2mm} \\
	The Chinese University of Hong Kong \\
	{\tt\small \{lirh,xzli,pheng,cwfu\}@cse.cuhk.edu.hk}
}

\newcommand{\TODO}[1]{{\color{red}{[TODO: #1]}}}
\newcommand{\phil}[1]{{\color[rgb]{0.3,0.7,0.3}{[PH: #1]}}}
\newcommand{\rh}[1]{{\color{black}{#1}}}
\newcommand{\xz}[1]{{\color[rgb]{1.0,0.6,0}{[XZ: #1]}}}
\newcommand{\dc}[1]{{\color{red}{[DC: #1]}}}
\newcommand{\para}[1]{\vspace{.05in}\noindent\textbf{#1}}
\def\ie{\emph{i.e.}}
\def\eg{\emph{e.g.}}
\def\etal{{\em et al.}}
\def\etc{{\em etc.}}

\maketitle
\thispagestyle{empty}

\begin{abstract}
We present PointAugment\footnote[1]{Code: \url{https://github.com/liruihui/PointAugment}}, a new auto-augmentation framework that automatically optimizes and augments point cloud samples to enrich the data diversity when we train a classification network. Different from existing auto-augmentation methods for 2D images, PointAugment is sample-aware and takes an adversarial learning strategy to jointly optimize an augmentor network and a classifier network, such that the augmentor can learn to produce augmented samples that best fit the classifier. Moreover, we formulate a learnable point augmentation function with a shape-wise transformation and a point-wise displacement, and carefully design loss functions to adopt the augmented samples based on the learning progress of the classifier. Extensive experiments also confirm PointAugment's effectiveness and robustness to improve the performance of various networks on shape classification and retrieval.

\end{abstract}
\vspace{-1mm}

\section{Introduction}
\label{sec:intro}

In recent years, there has been a growing interest in developing deep neural networks~\cite{qi2017pointnet,qi2017pointnet++,wang2018dynamic,mao2019interpolated,liu2019relation} for 3D point cloud processing.
%
%
To robustly train a network often relies on the availability and diversity of the data.
However, unlike 2D image benchmarks such as ImageNet~\cite{krizhevsky2012imagenet} and MS COCO dataset~\cite{lin2014microsoft}, which have over millions of training samples, 3D datasets are typically much smaller in quantity,
with relatively small amount of labels and limited diversity.
For instance, ModelNet40~\cite{wu20153d}, one of the most commonly-used benchmark for 3D point cloud classification, only has 12,311 models of 40 categories.
The limited data quantity and diversity may cause overfitting problem and further affect the generalization ability of the network.


Nowadays, data augmentation (DA) is a very common strategy to avoid overfitting and improve the network generalization ability by artificially enlarging the quantity and diversity of the training samples.
%
For 3D point clouds, due to the limited amount of training samples and an immense augmentation space in 3D, conventional DA strategies~\cite{qi2017pointnet,qi2017pointnet++}
often simply perturb the input point cloud randomly in a small and fixed pre-defined augmentation range to maintain the class label.
Despite its effectiveness for the existing classification networks, this conventional DA approach may lead to insufficient training, as summarized below.


\begin{figure}[!t]
\centering
\vspace*{-2mm}
\includegraphics[width=0.925\linewidth]{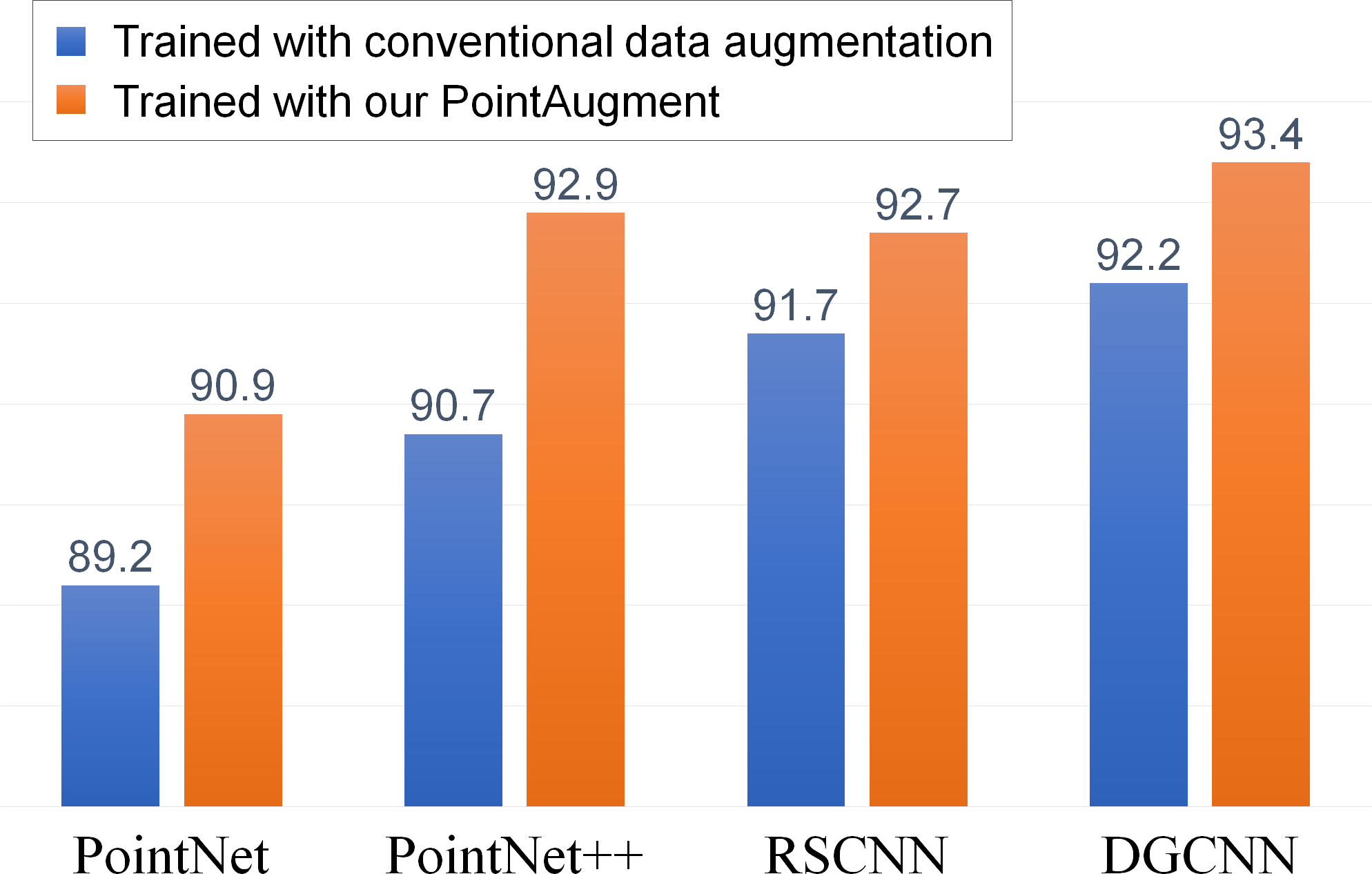}
\vspace*{-1mm}
\caption{Classification accuracy (\%) on ModelNet40 with or without training the networks with our PointAugment.
We can see clear improvements on four representative networks.
More comparison results are presented in Section~\ref{sec:experiment}.}
%
%
\label{fig:teaser}
\vspace*{-2mm}
\end{figure}

First, existing methods for deep 3D point cloud processing regard the network training and DA as two independent phases without {\em jointly optimizing\/} them,~\eg, feedback the training results to enhance the DA.
Hence, the trained network could be suboptimal.
%
Second, existing methods apply the {\em same fixed augmentation\/} process with rotation, scaling, and/or jittering, to all input point cloud samples.
The shape complexity of the samples is ignored in the augmentation,~\eg, a sphere remains the same no matter how we rotate it, but a complex shape may need larger rotations.
Hence, conventional DA may be redundant or insufficient for augmenting the training samples~\cite{graham2014fractional}.

To improve the augmentation of point cloud samples, we formulate \emph{PointAugment}, a new auto-augmentation framework for 3D point clouds, and demonstrate its effectiveness to enhance shape classification; see Figure~\ref{fig:teaser}.
Different from the previous works for 2D images, PointAugment {\em learns to produce augmentation functions specific to individual samples\/}.
Further, the learnable augmentation function considers both {\em shape-wise transformation\/} and {\em point-wise displacement\/}, which relate to the characteristics of 3D point cloud samples.
Also, PointAugment {\em jointly optimizes the augmentation process with the network training\/}, via an adversarial learning strategy to train the augmentation network (augmentor) together with the classification network (classifier) in an end-to-end manner.
By taking the classifier losses as {\em feedbacks\/}, the augmentor can learn to enrich the input samples by enlarging the intra-class data variations, while the classifier can learn to combat this by extracting insensitive features.
Benefited by such adversarial learning, the augmentor can then learn to generate augmented samples that best fit the classifier in different stages of the training, thus maximizing the capability of the classifier.

As the first attempt to explore auto-augmentation for 3D point clouds, we show by replacing conventional DA with PointAugment, clear improvements in shape classification on ModelNet40~\cite{wu20153d} (see Figure~\ref{fig:teaser}) and SHREC16~\cite{savva2016shrec16} (see Section~\ref{sec:experiment}) datasets can be achieved on four representative networks, including
PointNet~\cite{qi2017pointnet},
PointNet++~\cite{qi2017pointnet++},
RSCNN~\cite{liu2019relation}, and
DGCNN~\cite{wang2018dynamic}.
Also, we demonstrate the effectiveness of PointAugment on shape retrieval and evaluate its robustness, loss configuration, and modularization design.
More results are presented in Section~\ref{sec:experiment}.

\section{Related Work}
\label{sec:bg}


\para{Data augmentation on images.} \
Training data plays a very important role for deep neural networks to learn to perform tasks.
However, training data usually has limited quantity, compared with the complexity of our real world, so data augmentation is often needed as a means to enlarge the training set and maximize the knowledge that a network can learn from the training data.
Instead of randomly transforming the training data samples~\cite{zhang2018mixup,yun2019cutmix},
%
some works attempted to generate augmented samples from the original data by using image combination~\cite{lemley2017smart}, generative adversarial network (GAN)~\cite{shrivastava2017learning,ratner2017learning}, Bayesian optimization~\cite{tran2017bayesian}, and image interpolation in the latent space~\cite{devries2017dataset,liu2018feature,cheny2019multi}.
However, these methods may produce unreliable samples that are different from those in the original data.
On the other hand, some image DA techniques~\cite{lemley2017smart,zhang2018mixup,yun2019cutmix} apply pixel-by-pixel interpolation for images with regular structures;
however, they cannot handle order-invariant point clouds.

Another approach aims to find an optimal combination of predefined transformation functions to augment the training samples, instead of applying the transformation functions based on a manual design or by complete randomness.
AutoAugment~\cite{cubuk2019autoaugment} suggests a reinforcement learning strategy to find the best set of augmentation functions by alternatively training a proxy task and a policy controller, then applying the learned augmentation function to the input data.
Soon after, two other works, FastAugment~\cite{lim2019fast} and PBA~\cite{ho2019pba}, explore advanced hyper-parameter optimization methods to more efficiently find the best transformations for the augmentation.
%
Different from these methods, which learn to find a fixed augmentation strategy for all the training samples, PointAugment is sample-aware, meaning that we dynamically produce the transformation functions based on the properties of individual training samples and the network capability during the training process.

Very recently, Tang~\etal~\cite{tang2019ada} and Zhang~\etal~\cite{zhang2019adversarial} suggested to learn augmentation policies on target tasks using an adversarial strategy.
They tend to directly maximize the loss of augmented samples to improve the generalization of image classification networks.
Differently, PointAugment enlarges the loss between the augmented point clouds and their original ones by an explicitly-designed boundary (see Section~\ref{subsubsec:augmentor_loss} for details); it dynamically adjusts the difficulty of the augmented samples, so that the augmented samples can better fit the classifier for different training stages.

\para{Data augmentation on point cloud.}
In existing points processing networks, data augmentation mainly include random rotation about the gravity axis, random scaling, and random jittering~\cite{qi2017pointnet,qi2017pointnet++}.
These handcrafted rules are fixed throughout the training process, so we may not obtain the best samples to effectively train the network.
So far, we are not aware of any work that explores auto-augmentation to maximize the network learning with 3D point clouds.

%




\para{Deep learning on point cloud.} \
Improving on the PointNet architecture~\cite{qi2017pointnet}, several works~\cite{qi2017pointnet++,liu2019densepoint,liu2019relation} explored local structures to enhance the feature learning.
Some others explored the graph convolutional networks by creating a local graph~\cite{wang2018local,wang2018dynamic,shen2018mining,zhao2019pointweb} or geometric elements~\cite{landrieu2018large,prokudin2019efficient}.
Another stream of works~\cite{su2018splatnet,thomas2019kpconv,mao2019interpolated}
projected irregular points into a regular space to allow traditional convolutional neural networks to work on.
%
Different from the above works, our goal is not on designing a new network but on
boosting the classification performance of existing networks by effectively optimizing the augmentation of point cloud samples.
To this end, we design an augmentor to learn a sample-specific augmentation function and adjust the augmentation based also on the learning progress of the classifier.



\begin{figure}[!t]
\centering
\includegraphics[width=0.98\linewidth]{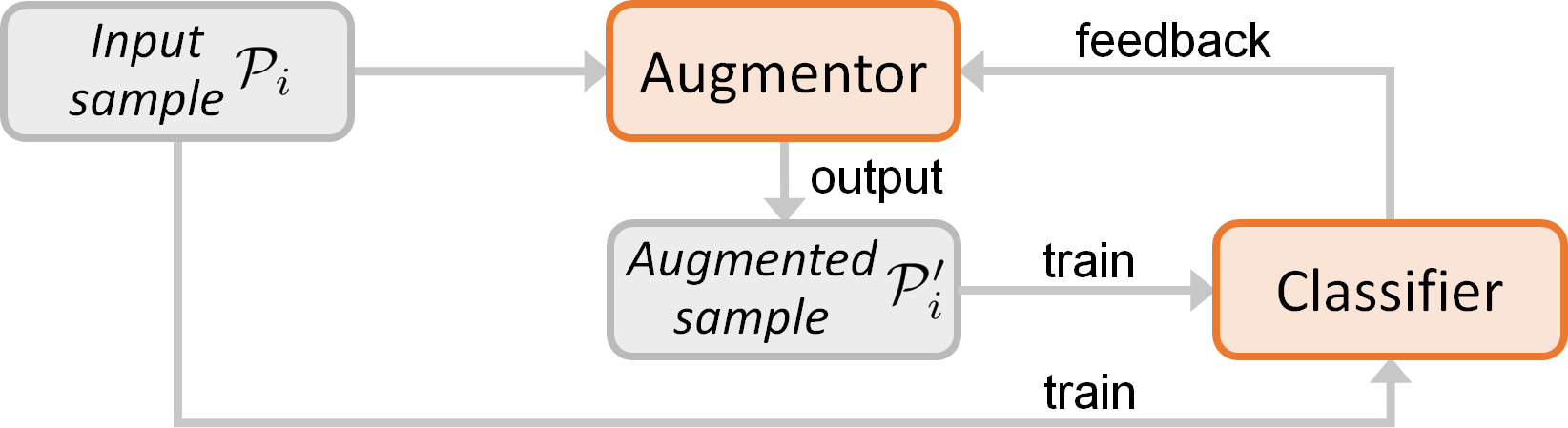}
\vspace*{-1mm}
\caption{An overview of our PointAugment framework.
We jointly optimize the augmentor and classifier
in an end-to-end manner with an adversarial learning strategy.}
\label{fig:overview}
\vspace*{-3.5mm}
\end{figure}

\section{Overview}
\label{sec:overview}

The main contribution of this work is the PointAugment framework that automatically optimizes the augmentation of the input point cloud samples for more effectively training the classification network.
Figure~\ref{fig:overview} illustrates the design of our framework, which has two deep neural network components:
(i) an augmentor $\mathcal{A}$ and (ii) a classifier $\mathcal{C}$.
Given an input training dataset $\{\mathcal{P}_i\}_{i=1}^M$ of $M$ samples, where each sample has $N$ points, before we train classifier $\mathcal{C}$ with sample $\mathcal{P}_i$, we feed $\mathcal{P}_i$ first to our augmentor $\mathcal{A}$ to generate an augmented sample $\mathcal{P}'_i$.
Then, we feed $\mathcal{P}_i$ and $\mathcal{P}'_i$ separately to classifier $\mathcal{C}$ for training, and further take $\mathcal{C}$'s results as feedback to guide the training of augmentor $\mathcal{A}$.

Before elaborating the PointAugment framework, we first discuss our key ideas behind the framework.
These are new ideas (not present in previous works~\cite{cubuk2019autoaugment,lim2019fast,ho2019pba}) that enable us to efficiently augment the training samples, which are now 3D point clouds instead of 2D images.

\begin{itemize}

\vspace*{-1.75mm}
\item
\emph{Sample-aware}.
Rather than finding a universal set of augmentation policy or procedure for processing every input data sample, we aim to regress a specific augmentation function for each input sample by considering the underlying geometric structure of the sample.
We call this a sample-aware auto-augmentation.
%

\vspace*{-1.25mm}
\item
\emph{2D vs. \hspace*{-2.5mm} 3D augmentation}.
Unlike 2D augmentations for images, 3D augmentation involves a more immense and different spatial domain.
Accounting for the nature of 3D point clouds, we consider two kinds of transformations on point cloud samples: shape-wise transformation (including rotation, scaling, and their combinations), and point-wise displacement (jittering of point locations), where our augmentor should learn to produce them to enhance the network training.
%

\vspace*{-1.25mm}
\item
\emph{Joint optimization}.
During the network training, the classifier will gradually learn and become more powerful, so we need more challenging augmented samples to better train the classifier, as the classifier becomes stronger.
Hence, we design and train the PointAugment framework in an end-to-end manner, such that we can jointly optimize both the augmentor and classifier.
To achieve so, we have to carefully design the loss functions and dynamically adjust the difficulty of the augmented samples, while considering both the input sample and the capacity of the classifier.




\end{itemize}

\section{Method}
\label{sec:method}

In this section, we first present the network architecture details of the augmentor and classifier (Section~\ref{subsec:network}).
Then, we present our loss functions formulated for the augmentor (Section~\ref{subsubsec:augmentor_loss}) and classifier (Section~\ref{subsubsec:cls_loss}), and introduce our end-to-end training strategy (Section~\ref{subsubsec:train_strategy}).
Lastly, we present the implementation details (Section~\ref{subsec:implementation}).


\subsection{Network Architecture}
\label{subsec:network}

\begin{figure}[!t]
\centering
\includegraphics[width=0.98\linewidth]{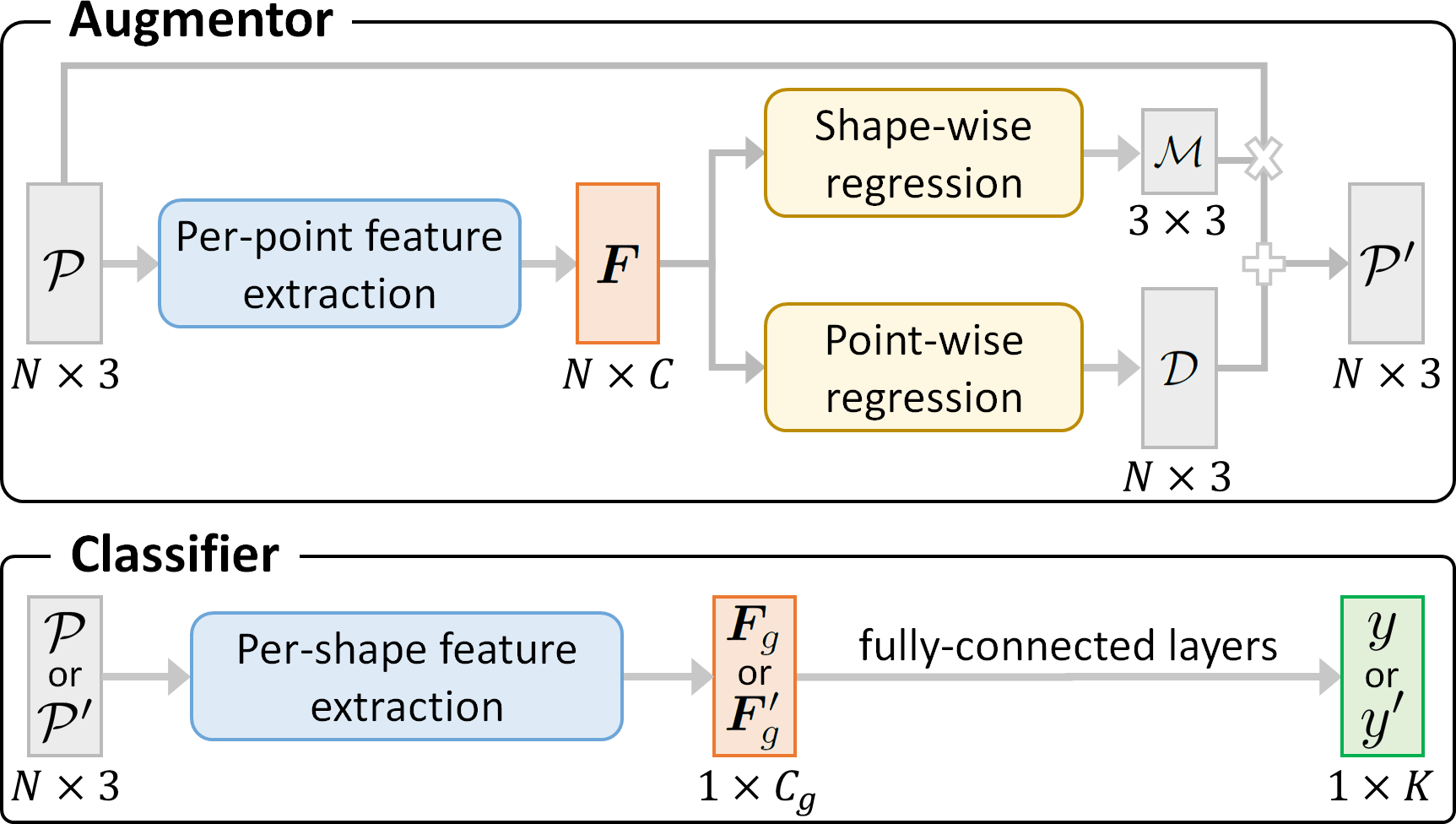}
\caption{Illustrations of the augmentor and classifier. The augmentor generates augmented sample $\mathcal{P}'$ from $\mathcal{P}$, and the classifier predicts the class label given $\mathcal{P}'$ or $\mathcal{P}$ as inputs.}
\label{fig:augmentor}
\end{figure}
\begin{figure}[!t]
\centering
\includegraphics[width=0.98\linewidth]{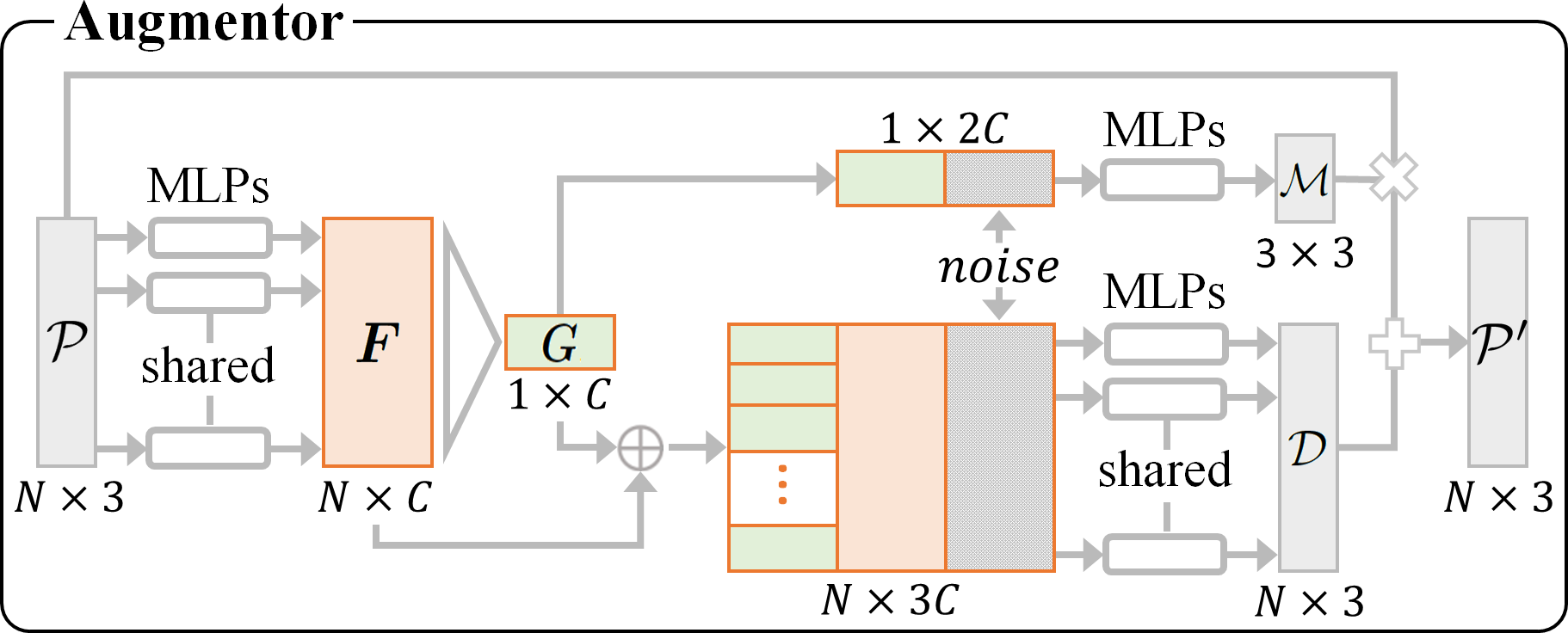}
\caption{Our implementation of the augmentor.}
\label{fig:aug_example}
\vspace*{-3mm}
\end{figure}

\para{Augmentor.} \
%
Different from existing works~\cite{cubuk2019autoaugment,lim2019fast,ho2019pba}, our augmentor is sample-aware, and it learns to generate a specific function for augmenting each input sample.
From now on, we drop subscript $i$ for ease of reading, and denote $\mathcal{P}$ as the training sample input to augmentor $\mathcal{A}$ and $\mathcal{P}'$ as the corresponding augmented sample output from $\mathcal{A}$.

The overall architecture of our augmentor is illustrated in Figure~\ref{fig:augmentor} (top).
First, we use a per-point feature extraction unit to embed point features $\bm{F} \in \mathbb{R}^{N \times C}$ for all $N$ points in $\mathcal{P}$, where $C$ is the number of feature channels.
From $\bm{F}$, we then regress the augmentation function specific to input sample $\mathcal{P}$ using two separate components in the architecture:
(i) shape-wise regression to produce transformation $\mathcal{M} \in \mathbb{R}^{3 \times 3}$ and
(ii) point-wise regression to produce displacement $\mathcal{D} \in \mathbb{R}^{N \times 3}$.
Note that, the learned $\mathcal{M}$ is a linear matrix in 3D space, combining mainly rotation and scaling, whereas the learned $\mathcal{D}$ gives point-wise translation and jittering.
Using $\mathcal{M}$ and $\mathcal{D}$, we can then generate the augmented sample $\mathcal{P}'$ as $\mathcal{P} \cdot \mathcal{M} + \mathcal{D}$.

The design of our proposed framework for the augmentor is generic, meaning that we may use different models to build its components.
%
Figure~\ref{fig:aug_example} shows our current implementation, for reference.
Specifically, similar to PointNet~\cite{qi2017pointnet}, we first employ a series of shared multi-layer perceptron (MLPs) to extract per-point features $\bm{F} \in \mathbb{R}^{N \times C}$.
We then employ max pooling to obtain the per-shape feature vector $\bm{G} \in \mathbb{R}^{1 \times C}$.
To regress $\mathcal{M}$, we generate a $C$-dimension noise vector based on a Gaussian distribution and concatenate it with $\bm{G}$, and then employ MLPs to obtain $\mathcal{M}$.
Note that the noise vector enables the augmentor to explore more diverse choices in regressing the transformation matrix through the randomness introduced into the regression process.
To regress $\mathcal{D}$, we concatenate $N$ copies of $\bm{G}$ with $\bm{F}$, together with an $N \times C$ noise matrix, whose values are randomly and independently generated based on a Gaussian distribution.
Lastly, we employ MLPs to obtain $\mathcal{D}$.

\para{Classifier.} \
Figure~\ref{fig:augmentor} (bottom) shows the general architecture of classifier $\mathcal{C}$.
It takes $\mathcal{P}$ and $\mathcal{P}'$ as inputs in two separate rounds and predicts corresponding class labels $y$ and $y'$.
Both $y$ and $y' \in \mathbb{R}^{1 \times K}$, where $K$ is the total number of classes in the classification problem.
In general, $\mathcal{C}$ first extracts per-shape global features $\bm{F}_g$ or $\bm{F}_g' \in \mathbb{R}^{1 \times C_g}$ (from $\mathcal{P}$ or $\mathcal{P}'$), and then employ fully-connected layers to regress a class label.
Also, the choice of implementing $\mathcal{C}$ is flexible.
We may employ different classification networks as $\mathcal{C}$.
In Section~\ref{sec:experiment}, we shall show that the performance of several conventional classification networks can be further boosted when equipped with our augmentor in the training.

\subsection{Augmentor loss}
\label{subsubsec:augmentor_loss}

To maximize the network learning, augmented sample $\mathcal{P}'$ generated by the augmentor should satisfy two requirements:
(i) $\mathcal{P}'$ should be more challenging than $\mathcal{P}$,~\ie, we aim for $L(\mathcal{P}') \geq L(\mathcal{P})$; and
(ii) $\mathcal{P}'$ should not lose its shape distinctiveness, meaning that it should describe a shape that is not too far (or different) from $\mathcal{P}$.

To achieve requirement (i), a simple way to formulate the loss function for the augmentor (denoted as $\mathcal{L}_{\mathcal{A}}$) is to maximize the difference between the cross entropy losses on $\mathcal{P}$ and $\mathcal{P}'$, or equivalently, to minimize
\begin{equation}
\label{equ:naive}
\mathcal{L}_{\mathcal{A}} = \exp[-(L(\mathcal{P}') - L(\mathcal{P}))],
\end{equation}
where
$L(\mathcal{P})$$=$$-\sum_{c=1}^{K}\hat{y}_c\log (y_c))$ is $\mathcal{P}$'s cross entropy loss;
$\hat{y}_c \in \{0,1\}$ denotes the one-hot ground-truth label when $\mathcal{P}$ belongs to the $c$-th class; and
$y_c \in [0,1]$ is the probability of predicting $\mathcal{P}$ as $c$-th class.
Note also that, for $\mathcal{P}'$ to be more challenging than $\mathcal{P}$, we assume that $L(\mathcal{P}')$$\geq$$ L(\mathcal{P})$ and a larger $L(\mathcal{P}')$ indicates a larger magnitude of augmentation, which can be defined as $\xi = L(\mathcal{P}')$$-$$L(\mathcal{P})$.


\begin{figure}[!t]
\centering
\includegraphics[width=0.65\linewidth]{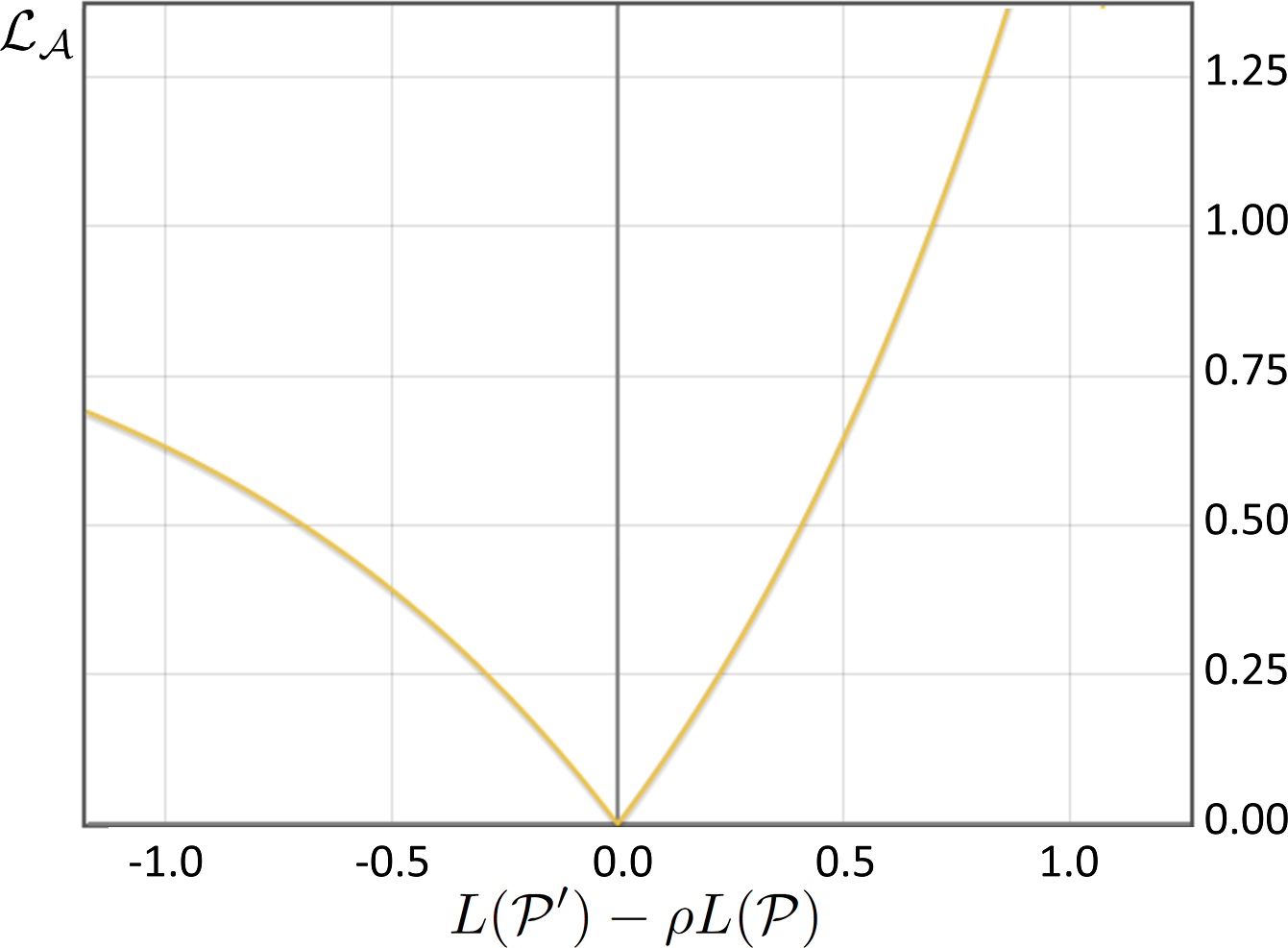}
\caption{Graph plot of Eq.~\eqref{equ:exp_margin}.}
\label{fig:aug_loss}
\vspace*{-1mm}
\end{figure}


However, if we naively minimize Eq.~\eqref{equ:naive} for $\mathcal{L}_{\mathcal{A}}$$\rightarrow$$0$, we encourage $L(\mathcal{P}')$$-$$L(\mathcal{P})$$\rightarrow$$\infty$.
So, a simple solution for $\mathcal{P}'$ is an arbitrary sample regardless of $\mathcal{P}$.
Such $\mathcal{P}'$ clearly violates requirement (ii).
Hence, we further restrict the augmentation magnitude $\xi$.
Inspired by LS-GAN~\cite{qi2017loss}, we first introduce a dynamic parameter $\rho$ and re-formulate $\mathcal{L}_{\mathcal{A}}$ as
\begin{equation}
\label{equ:exp_margin}
\mathcal{L}_{\mathcal{A}} = |1.0-\exp[L(\mathcal{P}') - \rho L(\mathcal{P})]|.
\end{equation}
See Figure~\ref{fig:aug_loss} for the graph plot of Eq.~\eqref{equ:exp_margin}.
In this formulation, we want $L(\mathcal{P}')$ to be large (for requirement (i)) but it should not be too large (for requirement (ii)), so we upper-bound $L(\mathcal{P}')$ by $\rho L(\mathcal{P})$.
Hence, we can obtain
\begin{equation}
\label{equ:margin}
\xi \ = \ L(\mathcal{P}') - L(\mathcal{P}) \ \leq \ (\rho-1) L(\mathcal{P}),
\end{equation}
where we denote $\xi_o = (\rho-1) L(\mathcal{P})$ as $\xi$'s upper bound.

Note that, when we train the augmentor, the classifier is fixed (to be presented in Section~\ref{subsubsec:train_strategy}), so $L(\mathcal{P})$ is fixed.
Hence, $\xi_o$ depends only on $\rho$.
Since it should be non-negative, we thus ensure $\rho$$\geq$$1$.
Moreover, considering that the classifier is very fragile at the beginning of the training, we pay more attention to training the classifier rather than generating a challenging $\mathcal{P}'$.
Hence, $\xi_o$ should not be too large, meaning that $\mathcal{P}'$ should not be too challenging.
Later, when the classifier becomes more powerful, we can gradually enlarge $\xi_o$ to allow the augmentor to generate a more challenging $\mathcal{P}'$.
Therefore, we design a dynamic $\rho$ to control $\xi_o$ with the following formulation:
\begin{equation}
\label{equ:rho}
\rho=\max\Big( 1 , \exp(\sum_{c=1}^{K}\hat{y}_c \cdot y_c) \Big),
\end{equation}
where $\max(1,*)$ ensures $\rho$$\geq$$1$.
At the beginning of the network training, the classifier predictions may not be accurate.
Hence, the prediction probability $y_c$ is generally small, resulting in a small $\rho$, and $\xi_o$ will also be small according to Eq.~\eqref{equ:margin}.
When the classifier becomes more powerful, $y_c$ will increase, and we will have larger $\rho$ and $\xi_o$ accordingly.


Lastly, to further ensure the augmented sample $\mathcal{P}'$ to be shape distinctive (for requirement (ii)), we add $L(\mathcal{P}')$, as a fidelity term, to Eq.~\eqref{equ:exp_margin} to construct the final loss $\mathcal{L}_{\mathcal{A}}$:
\begin{equation}
\label{equ:augmentor_loss}
	\mathcal{L}_{\mathcal{A}} = L(\mathcal{P}') + \lambda |1.0-\exp(L(\mathcal{P}') - \rho L(\mathcal{P}))|,
\end{equation}
where $\lambda$ is a fixed hyper-parameter to control the relative importance of each term.
A small $\lambda$ encourages the augmentor to focus more on the classification with less augmentation on $\mathcal{P}$, and vice versa.
In our implementation (all experiments), we set $\lambda=1$ to treat the two terms equally.

\subsection{Classifier loss}
\label{subsubsec:cls_loss}
The goal of the classifier $\mathcal{C}$ is to correctly predict both $\mathcal{P}$ and $\mathcal{P}'$.
Additionally, $\mathcal{C}$ should also have the ability to learn stable per-shape global features, no matter given $\mathcal{P}$ or $\mathcal{P}'$ as input.
We thus formulate the classifier loss $\mathcal{L}_{\mathcal{C}}$ as
\begin{equation}
\label{equ:cls_loss}
\mathcal{L}_{\mathcal{C}} = L(\mathcal{P}') + L(\mathcal{P}) + \gamma \|\bm{F}_g-\bm{F}_{g'}\|_2,
\end{equation}
where $\gamma$ is to balance the importance of the terms (we empirically set $\gamma$ as 10.0), and
$\|\bm{F}_g-\bm{F}_{g'}\|_2$ helps explicitly
penalize the feature difference between the augmented sample and the original one, and stabilize the network training.


\subsection{End-to-end training strategy}
\label{subsubsec:train_strategy}

Algorithm~\ref{alg:algorithm} summarizes our end-to-end training strategy.
Overall, the procedure alternatively optimizes and updates the learnable parameters in augmentor $\mathcal{A}$ and classifier $\mathcal{C}$, while fixing the other one, during the training.
Given input sample $\mathcal{P}_i$, we first employ $\mathcal{A}$ to generate its augmented sample $\mathcal{P}'_i$.
We then update the learnable parameters in $\mathcal{A}$ by calculating the augmentor loss using Eq.~\eqref{equ:augmentor_loss}.
In this step, we keep $\mathcal{C}$ unchanged.
After updating $\mathcal{A}$, we keep $\mathcal{A}$ unchanged, and generate the updated $\mathcal{P}'_i$.
We then feed $\mathcal{P}_i$ and $\mathcal{P}'_i$ to $\mathcal{C}$ one by one to obtain $L(\mathcal{P})$ and $L(\mathcal{P}')$, respectively, and update the learnable parameters in $\mathcal{C}$ by calculating the classifier loss using Eq.~\eqref{equ:cls_loss}.
In this way, we can optimize and train $\mathcal{A}$ and $\mathcal{C}$ in an end-to-end manner.



\subsection{Implementation details}
\label{subsec:implementation}

We implement PointAugment using PyTorch~\cite{paszke2017automatic}.
In detail, we set the number of training epochs $S=250$ with a batch size $B=24$.
To train the augmentor, we adopt the Adam optimizer with a learning rate of 0.001.
To train the classifier, we follow the respective original configuration from the released code and paper.
Specifically, for PointNet~\cite{qi2017pointnet}, PointNet++~\cite{qi2017pointnet++}, and RSCNN~\cite{liu2019relation}, we use the Adam optimizer with an initial learning rate of 0.001, which is gradually reduced with a decay rate of 0.5 every 20 epochs.
For DGCNN~\cite{wang2018dynamic}, we use the SGD solver with a momentum of 0.9 and a base learning rate of 0.1, which decays using a cosine annealing strategy~\cite{huang2017snapshot}.

Note also that, to reduce model oscillation~\cite{goodfellow2016nips}, we follow~\cite{shrivastava2017learning} to train PointAugment by using mixed training samples, which contain the original training samples as one half and our previously-augmented samples as the other half, rather than using only the original training samples.
Please refer to~\cite{shrivastava2017learning} for more details.
Moreover, to avoid overfitting, we set a dropout probability of 0.5 to randomly drop or keep the regressed shape-wise transformation and point-wise displacement.
In the testing phase, we follow previous networks~\cite{qi2017pointnet,qi2017pointnet++} to feed the input test samples to the trained classifier to obtain the predicted labels, without any additional computational cost.



\begin{algorithm}[!t]
\caption{Training Strategy in PointAugment}
\KwIn{training point sets $\{\mathcal{P}_i\}_{i=1}^M$,
corresponding ground-truth class labels $\{\hat{y}_i\}_{i=1}^M$, and
the number of training epochs $S$.}
\KwOut{$\mathcal{C}$ and $\mathcal{A}$.}
\For{$s=1,\cdots,S$}
{
    \For{$i=1,\cdots,M$}
    {
        \vspace*{3mm}
        \emph{// Update augmentor $\mathcal{A}$}\\
        Generate augmented sample $\mathcal{P}'_i$ from $\mathcal{P}_i$\\
        Calculate the augmentor loss using Eq.~\eqref{equ:augmentor_loss}\\
        Update the learnable parameters in $\mathcal{A}$\\
        \vspace*{3mm}
        \emph{// Update classifier $\mathcal{C}$}\\
        Calculate the classifier loss using Eq.~\eqref{equ:cls_loss} by feeding $\mathcal{P}_i$ and $\mathcal{P}'_i$ alternatively to $\mathcal{C}$\\
        Update the learnable parameters in $\mathcal{C}$\\
    }
}
\label{alg:algorithm}
\end{algorithm}

\section{Experiments}
\label{sec:experiment}

We conducted extensive experiments on PointAugment.
First, we introduce the benchmark datasets and classifiers employed in our experiments (Section~\ref{subsec:datasets}).
We then evaluate PointAugment on shape classification and shape retrieval (Section~\ref{subsec:evaluation}).
Next, we perform detailed analysis on PointAugment's robustness, loss configuration, and modularization design (Section~\ref{subsec:analysis}).
Lastly, we present further discussion and potential future extensions (Section~\ref{subsec:discussion}).


\subsection{Datasets and Classifiers}
\label{subsec:datasets}
\begin{table}
\caption{Statistics of the ModelNet10 (MN10)~\cite{wu20153d}, ModelNet40 (MN40)~\cite{wu20153d}, and SHREC16 (SR16)~\cite{savva2016shrec16} datasets, including the number of categories (classes), number of training and testing samples, average number of samples per class, and the corresponding standard deviation value.}
\label{tab:statistics}
\centering
\vspace{-5mm}
	\begin{center}
	\resizebox{0.99\linewidth}{!}{%
	\begin{tabular}{c|ccccc}
	\toprule[1pt]
	Dataset & \#Class & \#Training & \#Testing & Average & Std. \\ \hline \hline
	MN10 & 10 &3991 &908 & 399.10& 233.36\\
	MN40 & 40 &9843 &2468 & 246.07 & 188.64 \\
	SR16 & 55 &36148 &5165 & 657.22 & 1111.49 \\	
	\bottomrule[1pt]
	\end{tabular}}
	\end{center}
    \vspace{-3mm}
\end{table}

\para{Datasets.} \
We employed three 3D benchmark datasets in our evaluations,~\ie, ModelNet10~\cite{wu20153d}, ModelNet40~\cite{wu20153d}, and SHREC16~\cite{savva2016shrec16}, for which we denote as MN10, MN40, and SR16, respectively.
Table~\ref{tab:statistics} presents statistics about the datasets,
%
showing that, MN10 is a very small dataset with only 10 classes.
Though most networks~\cite{qi2017pointnet,liu2019densepoint} can achieve a high classification accuracy on MN10, they may easily overfit.
SR16 is the largest data with over 36,000 training samples.
However, the high standard deviation (std.) value,~\ie, 1111, shows the uneven distribution of training samples among the classes.
For example, in SR16, the \emph{Table} class has 5,905 training samples, while the \emph{Cap} class has only 39 training samples.
For MN40, we directly adopt the data kindly provided by PointNet~\cite{qi2017pointnet} and follow the same train-test split.
For MN10 and SR16, we uniformly sample 1,024 points on each mesh surface and normalize the point sets to fit a unit ball centered at the origin.

\para{Classifiers.} \
As explained in Section~\ref{subsec:network}, our overall framework is generic, and we can employ different classification networks as classifier $\mathcal{C}$.
To show that the performance of conventional classification networks can be further boosted when equipped with our augmentor, in the following experiments, we employ several representative classification networks as classifier $\mathcal{C}$, including
(i)   PointNet~\cite{qi2017pointnet}, a pioneer network that processes points individually;
(ii)  PointNet++~\cite{qi2017pointnet++}, a hierarchical feature extraction network;
(iii) RSCNN\footnote{Only the single-scale RSCNN~\cite{liu2019relation} is released so far.}~\cite{liu2019relation}, a recently-released enhanced version of PointNet++ with a relation weight inside each local region; and
(iv)  DGCNN~\cite{wang2018dynamic}, a graph-based feature extraction network.
Note that, most existing networks~\cite{zhang2019shellnet,thomas2019kpconv,liu2019densepoint} are built and extended from the above networks with various means of adaptation.

\newcommand{\BE}[1]{{\textbf{#1}}}
\begin{table}
	\caption{Comparing the overall shape classification accuracy (\%) on MN40, MN10, and SR16, for various classifiers equipped with conventional DA (first four rows) and with our PA (last four rows); PA denotes PointAugment.
	We can observe improvements for all datasets and all classifiers.}
	\label{tab:classification}
	\centering
    \vspace{-5mm}
	\begin{center}
		\resizebox{0.925\linewidth}{!}{%
			\begin{tabular}{c|ccc}
				\toprule[1pt]
				Method & MN40 & MN10 & SR16 \\ \hline \hline
				PointNet~\cite{qi2017pointnet} &89.2 &91.9 & 84.4 \\
				PointNet++~\cite{qi2017pointnet++} &90.7 &93.3 & 85.1\\
                		RSCNN~\cite{liu2019relation} &91.7 &94.2  & 86.6 \\
				DGCNN~\cite{wang2018dynamic} &92.2 &94.8 & 87.0 \\	
                	\hline \hline
				PointNet (+PA) &90.9 (1.7$\uparrow$) &94.1 (2.2$\uparrow$) & 88.4 (4.0$\uparrow$) \\
                		PointNet++ (+PA) &92.9 (2.2$\uparrow$) &95.8 (2.5$\uparrow$) & 89.5  (4.4$\uparrow$) \\
                		RSCNN (+PA) &92.7 (1.0$\uparrow$) &96.0 (1.8$\uparrow$)  & 90.1 (3.5$\uparrow$) \\
                		DGCNN (+PA) &93.4 (1.2$\uparrow$) &96.7 (1.9$\uparrow$) & 90.6 (3.6$\uparrow$) \\
				\bottomrule[1pt]
		    \end{tabular}}
	\end{center}
\vspace{-4mm}
\end{table}


\subsection{PointAugment Evaluation}
\label{subsec:evaluation}

\para{Shape classification}.
First, we evaluate our PointAugment on the shape classification task using the classifiers listed in Section~\ref{subsec:datasets}.
For comparison, when we train the classifiers without PointAugment, we follow~\cite{qi2017pointnet++} to augment the training samples by random rotation, scaling, and jittering, which are considered as conventional DA.

Table~\ref{tab:classification} summarizes the quantitative evaluation results for comparison.
We report the overall classification accuracy (\%) of each classifier on all the three benchmark datasets, with conventional DA and with our PointAugment.
From the results we can clearly see that, by employing PointAugment, the shape classification accuracies of all classifier networks can improve for all the three benchmark datasets.
Particularly, on MN40, the classification accuracy achieved by DGCNN+PointAugment is 93.4\%, which is a very high accuracy value comparable with the very recent works~\cite{zhang2019shellnet,thomas2019kpconv,liu2019densepoint}.
Moreover, our PointAugment is shown to be more effective on the imbalanced SR16 dataset; see the right-most column in Table~\ref{tab:classification}, showing that PointAugment can alleviate the class size imbalance problem through our sample-aware auto-augmentation strategy to introduce more intra-class variation to the augmented samples.


\begin{table}
	\caption{Comparing the shape retrieval results (mAP, \%) on MN40, for various methods equipped with conventional DA or with our PointAugment.
	Again, we can observe clear improvements in retrieval accuracy for all the four methods.}
	\label{tab:retrieval}
	\centering
    \vspace{-5mm}
	\begin{center}
		\resizebox{0.99\linewidth}{!}{%
			\begin{tabular}{c|ccc}
				\toprule[1pt]
				Method &  Conventional DA & PointAugment & Change \\ \hline \hline
				PointNet~\cite{qi2017pointnet}		&70.5 &75.8 & 5.2$\uparrow$ \\
				PointNet++~\cite{qi2017pointnet++}	&81.3 &86.7 & 5.4$\uparrow$ \\
                RSCNN~\cite{liu2019relation} 		&83.2 &86.6 & 3.4$\uparrow$ \\
				DGCNN~\cite{wang2018dynamic} 		&85.3 &89.0 & 3.7$\uparrow$ \\	
				\bottomrule[1pt]
		    \end{tabular}}
	\end{center}
\vspace{-2mm}
\end{table}

\begin{figure}
	\centering
	\includegraphics[width=0.95\linewidth]{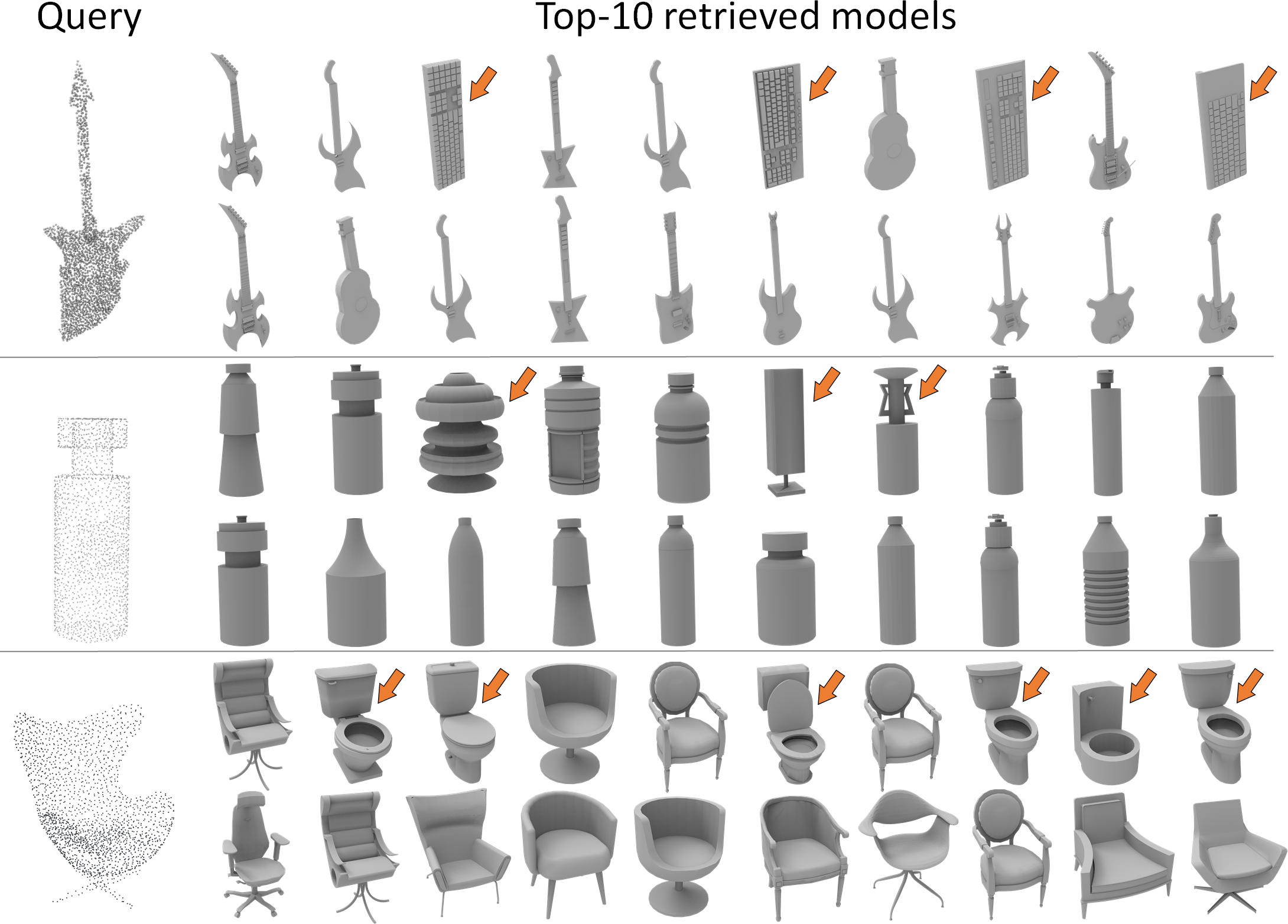}
	\vspace{-1mm}
	\caption{Shape retrieval results on MN40.
	For each query shape on the left, we present two rows of Top-10 retrieval results: the top row uses PointNet~\cite{qi2017pointnet} and the bottom row uses PointNet+PointAugment.
	Note that the obviously-wrong retrieval results are marked with red arrows.}
	\label{fig:retrieval}
	\vspace{-2mm}
\end{figure}

\para{Shape retrieval}.
To validate whether PointAugment facilitates the classifiers to learn a better shape signature, we compare the shape retrieval performance on MN40.
Specifically, we regard each sample in the testing split as a query, and aim to retrieve the best similar shapes from the testing split by comparing the cosine similarity between their global features $\bm{F}_g$.
In this experiment, we employ the mean Average Precision (mAP) as the evaluation metric.

Table~\ref{tab:retrieval} presents the evaluation results, which clearly show that PointAugment improves the shape retrieval performance for all the four classifier networks.
Especially, for PointNet~\cite{qi2017pointnet} and PointNet++~\cite{qi2017pointnet++}, the percentage of improvement is over 5\%.
Besides, we show visual results on shape retrieval for three different query models in Figure~\ref{fig:retrieval}.
Compared with the original PointNet~\cite{qi2017pointnet}, which is equipped with conventional DA, the augmented version with PointAugment produces more accurate retrievals.


\subsection{PointAugment Analysis}
\label{subsec:analysis}

Further, we conducted more experiments to evaluate various aspects of PointAugment, including
a robustness test (Section~\ref{subsubsec:robustness}),
an ablation study (Section~\ref{subsubsec:ablation}), and
a detailed analysis on its Augmentor network (Section~\ref{subsubsec:augmentor_analysis}).
Note that, in these experiments, we employ PointNet++~\cite{qi2017pointnet++} as the classifier and perform experiments on MN40.


\subsubsection{Robustness Test}
\label{subsubsec:robustness}
\begin{table}
	\caption{Robustness test to compare our PointAugment with conventional DA.
	Here, we corrupt each input test sample
	by random jittering (Jitt.) with Gaussian noise in [-1.0, 1.0],
	by scaling with a ratio of 0.9 or 1.1, or
	by a rotation of $90^{\circ}$ or $180^{\circ}$ along the gravity axis.
	Also, we show the original accuracy (Ori.) without using corrupted samples.}
	\label{tab:robustness}
	\centering
    \vspace{-3mm}
	\begin{center}
		\resizebox{0.99\linewidth}{!}{%
			\begin{tabular}{c|c|c|c|c|c|c}
				\toprule[1pt]
				Method  & Ori. & Jitt. & 0.9 & 1.1 & $90^{\circ}$ & $180^{\circ}$  \\ \hline \hline
\rh{Without DA} &89.1 & 88.2 & 88.2 & 88.2 & 48.2 & 40.1  \\
Conventional DA &90.7 & 90.3 & 90.3 & 90.3 & 89.9 & 89.7  \\
                PointAugment  &\textbf{92.9} & \textbf{92.8} & \textbf{92.8} & \textbf{92.8} & \textbf{92.7} & \textbf{92.6}  \\
				\bottomrule[1pt]
		    \end{tabular}}
	\end{center}
\vspace{-2mm}
\end{table}


\begin{table}
  \centering
  \caption{Ablation study of PointAugment. $\mathcal{D}$: point-wise displacement, $\mathcal{M}$: shape-wise transformation, DP: dropout, and Mix: mixed training samples (see Section~\ref{subsubsec:train_strategy}).}
  \label{tab:ablation}
  \resizebox{0.85\linewidth}{!}{%
  \begin{tabular}{c|ccccc|cc}
    \toprule[1pt]
    Model & $\mathcal{D}$& $\mathcal{M}$ & DP & Mix & Acc. & Inc.$\uparrow$\\
    \hline \hline
    A &  & &  &   & 90.7 &  - \\
    B &  \checkmark&  &    &  &91.7  &  1.0\\
    C &  & \checkmark  &    &  &91.9  &  1.2\\
    D &  \checkmark&  \checkmark &  &    &92.5&  1.8  \\
    E &  \checkmark&  \checkmark & \checkmark &    &92.8&  2.1  \\
    F &  \checkmark&  \checkmark &  \checkmark& \checkmark   &\textbf{92.9}&  \textbf{2.2}  \\
    \bottomrule[1pt]
  \end{tabular}}
  \vspace{-1mm}
\end{table}

We conducted the robustness test by corrupting test samples using the following five settings:
(i) adding random jittering with Gaussian noise ranged $[-1.0, 1.0]$;
(ii,iii) adding uniform scaling with a ratio of 0.9 or 1.1; and
(iv,v) adding rotation with $90^{\circ}$ or $180^{\circ}$ along the gravity axis.
\rh{For each setting, we use
three different DA strategies: without DA, conventional DA, and our PointAugment.}

Table~\ref{tab:robustness} reports the results, where we show also the original test accuracy (Ori.) without using corrupted test samples as a reference.
\rh{The results in the first two rows show that DA is an efficient way to improve the classification performance. Further,}
comparing the \rh{last} two rows in the table, we can see that for all settings, our PointAugment consistently outperforms the conventional DA, \rh{which is random-based and may not yield good augmented samples all the time}.
Particularly, by comparing the results with the original test accuracy, PointAugment is less sensitive to corruption, where the achieved accuracy reduces only slightly.
Such a result shows that PointAugment improves the robustness of a network with better shape recognition.


\begin{figure}
	\centering
	\includegraphics[width=0.98\linewidth]{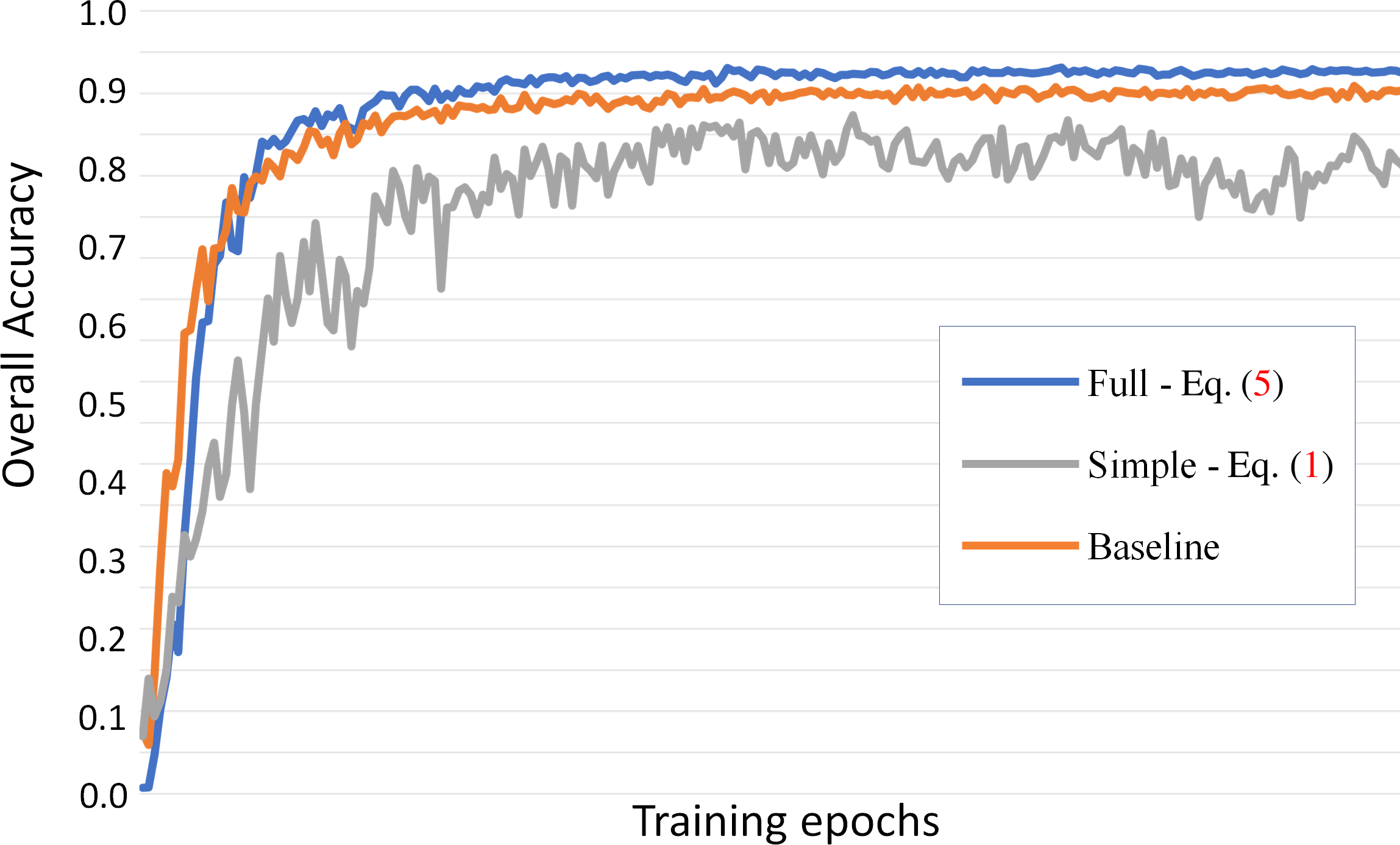}
	\caption{Evaluation curves: shape classification accuracy using different versions of $\mathcal{L}_{\mathcal{A}}$ over training epochs.}
	\label{fig:loss_analysis}
	\vspace{-2mm}
\end{figure}


\vspace{-1mm}
\subsubsection{Ablation Study}
\label{subsubsec:ablation}

Table~\ref{tab:ablation} summarizes the results of the ablation study.
Model A denotes PointNet++~\cite{qi2017pointnet++} without our augmentor, which gives a baseline classification accuracy of 90.7\%.
On top of Model A, we employ our augmentor with point-wise displacement $\mathcal{D}$ alone (Model B), with shape-wise transformation $\mathcal{M}$ alone (Model C), or with both (Model D).
From the results shown in the first four rows in Table~\ref{tab:ablation}, we can see that, each of the augmentation functions contributes to produce more effective augmented samples.

Besides, we also ablate the dropout strategy (DP) for training, and the use of mixed training samples (Mix), as presented in Section~\ref{subsec:implementation}, where we create Models E \& F for comparison; see Table~\ref{tab:ablation}.
By comparing the classification accuracies achieved by Models D, E, and F, we can see that both DP and Mix help to slightly improve the overall results. Note, these strategies are typically for stabilizing the model
training and exploring more transformations.



\vspace{-2mm}
\subsubsection{Augmentor Analysis}
\label{subsubsec:augmentor_analysis}

\vspace*{-1mm}
\para{Analysis on $\mathcal{L}_{\mathcal{A}}$.} \
As described in Section~\ref{subsubsec:augmentor_loss}, we employ $\mathcal{L}_{\mathcal{A}}$ (see Eq.~\eqref{equ:augmentor_loss}) to guide the training of our augmentor.
To demonstrate its superiority, we compare it with (i)
a simple version (see Eq.~\eqref{equ:naive}) and (ii) a baseline,~\ie, the conventional DA employed in PointNet++~\cite{qi2017pointnet++}.
Figure~\ref{fig:loss_analysis} plots the \rh{evaluation} accuracy curves in terms of the training epochs.

Clearly, the training state achieved by using the simple version is very unstable; see the gray plot.
This indicates that simply enlarging the difference between $\mathcal{P}$ and $\mathcal{P}'$ without restriction will create turbulence in the training process, resulting in a worse classification performance, when compared even with the baseline; see the orange plot.
Comparing the blue and orange plots in Figure~\ref{fig:loss_analysis}, we can see that, at the beginning of the training, since the augmentor is initialized randomly, the accuracy of employing PointAugment is slightly lower than the baseline.
However, when the training continues, PointAugment rapidly surpasses the baseline and shows a clear improvement over the baseline, showing the effectiveness of our designed augmentor loss.

\begin{table}[!t]
	\caption{Classification accuracy (\%) for diff. $\lambda$ in Eq.~\eqref{equ:augmentor_loss}.}
	\label{tab:selection}
	\centering
    \vspace{-5mm}
	\begin{center}
		\resizebox{0.55\linewidth}{!}{%
			\begin{tabular}{c|c|c}
				\toprule[1pt]
				$\lambda=0.5$  & $\lambda=1.0$ & $\lambda=2.0$ \\ \hline 
				92.1 &\textbf{92.9} & 92.3 \\
				\bottomrule[1pt]
		    \end{tabular}}
	\end{center}
    \vspace{-4mm}
\end{table}

\begin{table}[!t]
	\caption{Accuracy (\%) with diff. feature extraction units.}
	\label{tab:extraction}
	\centering
    \vspace{-5mm}
	\begin{center}
		\resizebox{0.55\linewidth}{!}{%
			\begin{tabular}{c|c|c}
				\toprule[1pt]
				DenseConv & EdgeConv & our \\ \hline 
				92.5  & 92.7 & \textbf{92.9} \\
				\bottomrule[1pt]
		    \end{tabular}}
	\end{center}
    \vspace*{-6mm}
\end{table}

\para{Hyper-parameter $\lambda$ in Eq.~\eqref{equ:augmentor_loss}}. \
Table~\ref{tab:selection} shows the classification accuracy for different choices of $\lambda$.
As we mentioned in Section~\ref{subsubsec:augmentor_loss}, a small $\lambda$ encourages the augmentor to focus more on the classification.
However, if it is too small,~\eg, 0.5, the augmentor tends to take no actions in the augmentation function, thus leading to a worse classification performance; see the comparisons in the left two columns in Table~\ref{tab:selection}.
On the other hand, if we set a larger $\lambda$,~\eg, 2.0, the augmented samples may be too difficult for the classifier.
Such a result also hinders the network training; see the right-most column.
Hence, in PointAugment, we adopt $\lambda=1.0$ to equally treat the two components.


\para{Analysis on the augmentor architecture design.} \
As we mentioned in Section~\ref{subsec:network}, we employ a per-point feature extraction unit to embed per-point features $\bm{F}$ given the training sample $\mathcal{P}$.
In our implementation, we use shared MLPs to extract $\bm{F}$.
In this part, we further explore two other choices of feature extraction units for replacing the MLPs, including DenseConv~\cite{liu2019densepoint} and EdgeConv~\cite{wang2018dynamic}.
Please refer to their original papers for the detailed methods.
Table~\ref{tab:extraction} shows the accuracy comparisons for the three implementations.
From the results, we can see that, although using MLPs is a relative simple implementation compared with DenseConv and EdgeConv, it can lead to the best classification performance.
We think of the following reasons.
The aim of our augmentor is to regress a shape-wise transformation $\mathcal{M} \in \mathbb{R}^{3 \times 3}$ and a point-wise displacement $\mathcal{D} \in \mathbb{R}^{N \times 3}$ from the per-point features $\bm{F}$, which is not a very tough task.
If we apply a complex unit to extract $\bm{F}$, it may easily have overfitting problem.
The results shown in Table~\ref{tab:extraction} demonstrate that the MLPs is already enough for our augmentor to regress the augmentation functions.





\subsection{Discussion and Future work}
\label{subsec:discussion}


Overall, the augmentor network learns the sample-level augmentation function in
a self-supervised manner, by taking the feedback from the classifier to update its parameters.
As a result, the advantage on the classifier network is that by exploring those well-tuned augmented samples, the classifier can enhance its capability and better learn to uncover intrinsic variations among the different classes and discover the intra-class insensitive features.

In the future, we plan to adapt PointAugment for more tasks, such as part segmentation~\cite{qi2017pointnet,liu2019relation}, semantic segmentation~\cite{meng2019vv,zhao2019pointweb,chen2020lassonet}, object detection~\cite{yang2019std,shi2019pointrcnn}, upsampling~\cite{yu2018pu, li2019pu}, denoising~\cite{hermosilla2019total,rakotosaona2019pointcleannet}, etc.
However, it is worth to note particularly that different tasks require different considerations.
For example, for parts segmentation, it is better for the augmentor to be part-aware and produce distortions on parts without changing the parts semantic; for object detection, the augmentor should be able to generate a richer variety of 3D transformations for various kinds of object instances in 3D scenes, etc.
%
%
Therefore, one future direction is to explore \rh{part-aware or} instance-aware auto-augmentation to extend PointAugment for other tasks.

\section{Conclusion}
\label{sec:conclusion}

We presented PointAugment, the first auto-augmentation framework that we are aware of for 3D point clouds, considering both the capability of the classification network and the complexity of the training samples.
%
First, PointAugment is an end-to-end framework
that jointly optimizes the augmentor and classifier networks, such that the augmentor can learn to improve based on feedback from the classifier and the classifier can learn to process wider variety of training samples.
%
Second, PointAugment is sample-aware with its augmentor learns to produce augmentation functions specific to the input samples, with a shape-wise transformation and a point-wise displacement for handling point cloud samples.
Third, we formulate a novel loss function to enable the augmentor to dynamically adjust the augmentation magnitude based on the learning state of the classifier, so that it can generate augmented samples that best fit the classifier in different training stages.
%
In the end, we conducted extensive experiments and demonstrated how PointAugment contributes to improve the performance of four representative networks on the MN40 and SR16 datasets.

\para{Acknowledgments.} \
We thank anonymous reviewers for the valuable comments.
The work is supported by the Research Grants Council of the Hong Kong Special Administrative Region (CUHK 14201717 \& 14201918), and CUHK Research Committee Direct Grant for Research 2018/19.

{\small
\bibliographystyle{ieee_fullname}
\bibliography{egbib}
}

\end{document}